\documentclass[final]{cvpr}
\usepackage{times}
\usepackage{epsfig}
\usepackage{graphicx}
\usepackage{amsmath}
\usepackage{amssymb}
\usepackage{amsthm}
\usepackage{siunitx}
\usepackage[export]{adjustbox}
\usepackage[linesnumbered,ruled,vlined]{algorithm2e}
\usepackage{float}
\usepackage{algpseudocode}
\usepackage{multirow}

\usepackage[pagebackref=true,breaklinks=true,colorlinks,bookmarks=false]{hyperref}

\begin{document}

\title{PCA Event-Based Optical Flow for Visual Odometry}

\author{Mahmoud Z. Khairallah
\hspace{1cm}
Fabien Bonardi
\hspace{1cm}
David Roussel
\hspace{1cm}
Samia Bouchafa\\
IBISC, Univ. Evry, Université Paris-Saclay, 91025, Evry, France\\
{\tt\small \{mahmoud.khairallah, fabien.bonardi, david.roussel, samia.bouchafa\}@univ-evry.fr}}

\maketitle
\begin{abstract}
With the advent of neuromorphic vision sensors such as event-based cameras,  a paradigm shift is required for most computer vision algorithms. Among these algorithms, optical flow estimation is a prime candidate for this process considering that it is linked to a neuromorphic vision approach. Usage of optical flow is widespread in robotics applications due to its richness and accuracy.  We present a  Principal Component Analysis (PCA) approach to the problem of event-based optical flow estimation. In this approach, we examine different regularization methods which efficiently enhance the estimation of the optical flow. We show that the best variant of our proposed method, dedicated to the real-time context of visual odometry, is about two times faster compared to state-of-the-art implementations while significantly improves optical flow accuracy.

\end{abstract}

\section{Introduction}
Neuromorphic cameras are considered to be one of the revolutionary visual sensors that have been developed over the past decade\cite{survey}. Based on the demand for robotics research to overcome all the known problems caused by standard cameras (i.e. image blur, low dynamic range and data redundancy), neuromorphic sensors are still currently being developed to overcome most of the standard cameras' drawbacks. 
Instead of transmitting a sequence of whole images at a constant rate, which includes a lot of redundant information and can cause high latency, the pixels of these neuromorphic chips operate independently and asynchronously.\\
\begin{figure}[t]
\begin{center}
\fbox{\includegraphics[width=0.75\linewidth]{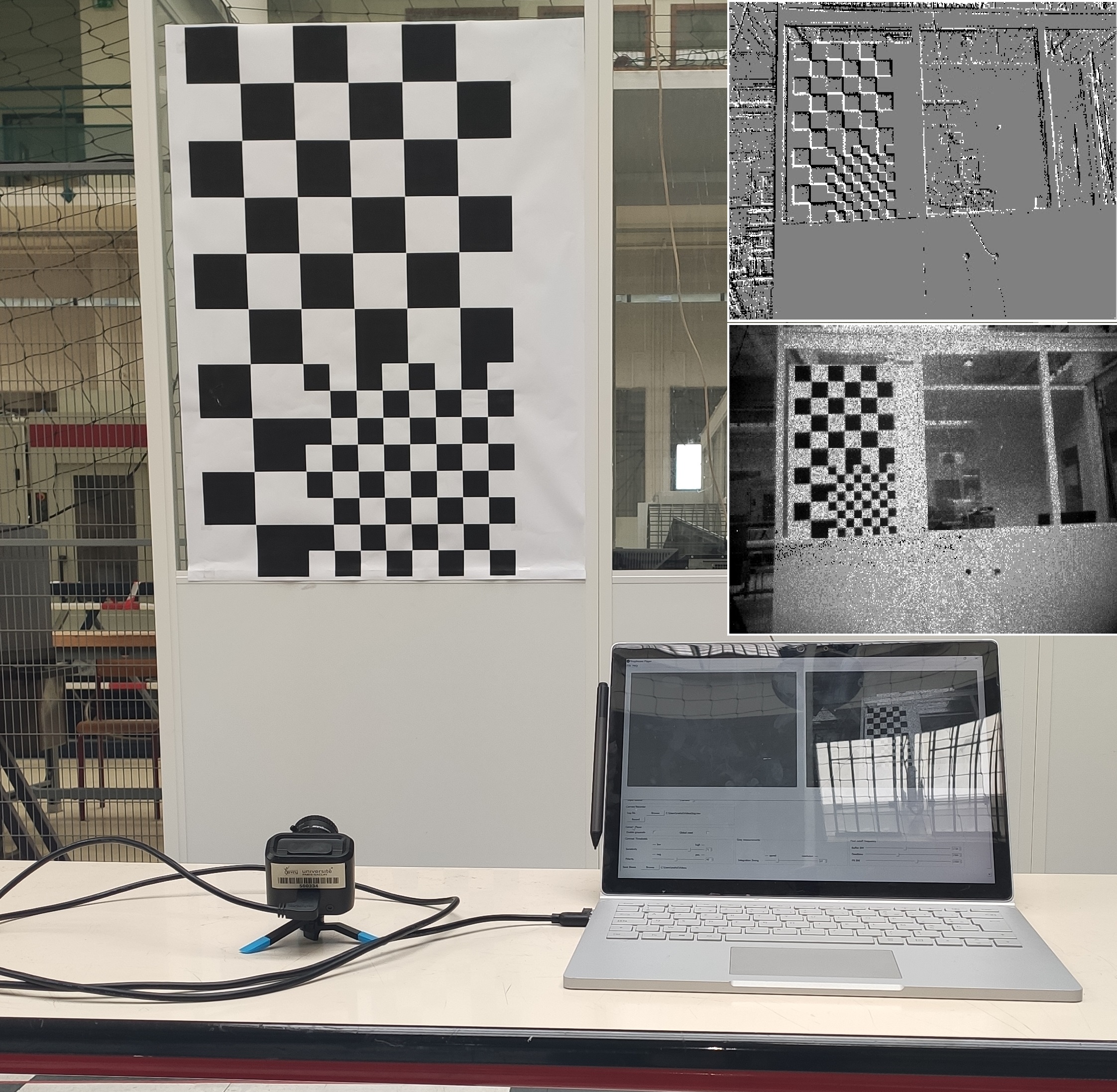}}
    
\end{center}
   \caption{The setting of our environment where the checkerboard is tracked by ATIS camera. Right: the change detection output obtained by the camera shown above the grayscale output}
\label{fig:davis}
\end{figure}
\noindent
These sensors can provide two kinds of information: the differentiation of light intensity of the environment as a raw output and the integration of this output to recover grayscale information (see Figure \ref{fig:davis}). 
Due to the independence of each pixel, the raw output of a neuromorphic sensor can be provided with a very low temporal resolution that may be less than 1 $\mu s$. This high-debit datastream solves the problem of motion blur and extends the dynamic range to more than $120$ dB.\\

\noindent
Considering events in the design of new algorithms naturally suggests fully exploiting neuromorphic sensors for a complete and precise estimation of optical flow. Indeed, events are due to changes in the environment and therefore indirectly linked to optical flow information. Moreover, optical flow is essential in many applications like robotics since it can be used as a core stage for many tasks (visual odometry or SLAM) that require accurate and fast optical flow algorithms. 
In this paper, we show improvements in event-based optical flow estimation using a PCA (Principal Component Analysis) scheme and explore the availability of using our algorithm in real-time applications.

\section{Related Work}
The events generated by neuromorphic vision sensors carry intrinsically optical flow information because they respond only to changes in luminosity. 
The asynchronous quasi-continuous nature of these sensors promotes their incorporation in critical robotics applications which require agility and accuracy. 
As one of the first attempts to tackle the challenge of proposing an event-driven optical flow method, Delbruck \cite{delbruck8}, considers it as a problem of data association since each event is created when gradients change. 
Events are correlated using the difference of timestamps between a considered event and the events in its spatio-temporal neighborhood. Hence, the edge that created the event is assigned one of four directions from $\si{0\degree}$ to $\si{135\degree}$ separated by $\si{45\degree}$. 
The final direction and magnitude of the event's optical flow are determined using the time of flight between the events stacked perpendicular to the edge direction. 
This method can be effective to augment the amount of information induced by each event but can not be considered as a suitable method of optical flow estimation because it gives highly discretized directions and a noisy estimated magnitude.\\

\noindent
Since each event is the differential log output of light intensity, Benosman et al \cite{benosman12}, integrate (stack) the events in the event spatio-temporal neighborhood and use the count of events at each pixel as an intensity-equivalent value. 
The count of events is inserted in an adapted event-based Lucas-Kanade scheme to estimate the optical flow. 
This method can be considered as the first event-based optical flow algorithm. However, the intensity-equivalent value is not the real intensity value. As a consequence, the optical flow magnitude may not correspond accurately to the real magnitude. 
Moreover, to get sufficiently accurate estimations of optical flow, the value of $\Delta t$ that is related to the size of the spatio-temporal neighborhood needs to be tuned each time for different dynamics of the environment.\\
\begin{figure}[t]
    \begin{center}
        \fbox{\includegraphics[width=0.9\linewidth]{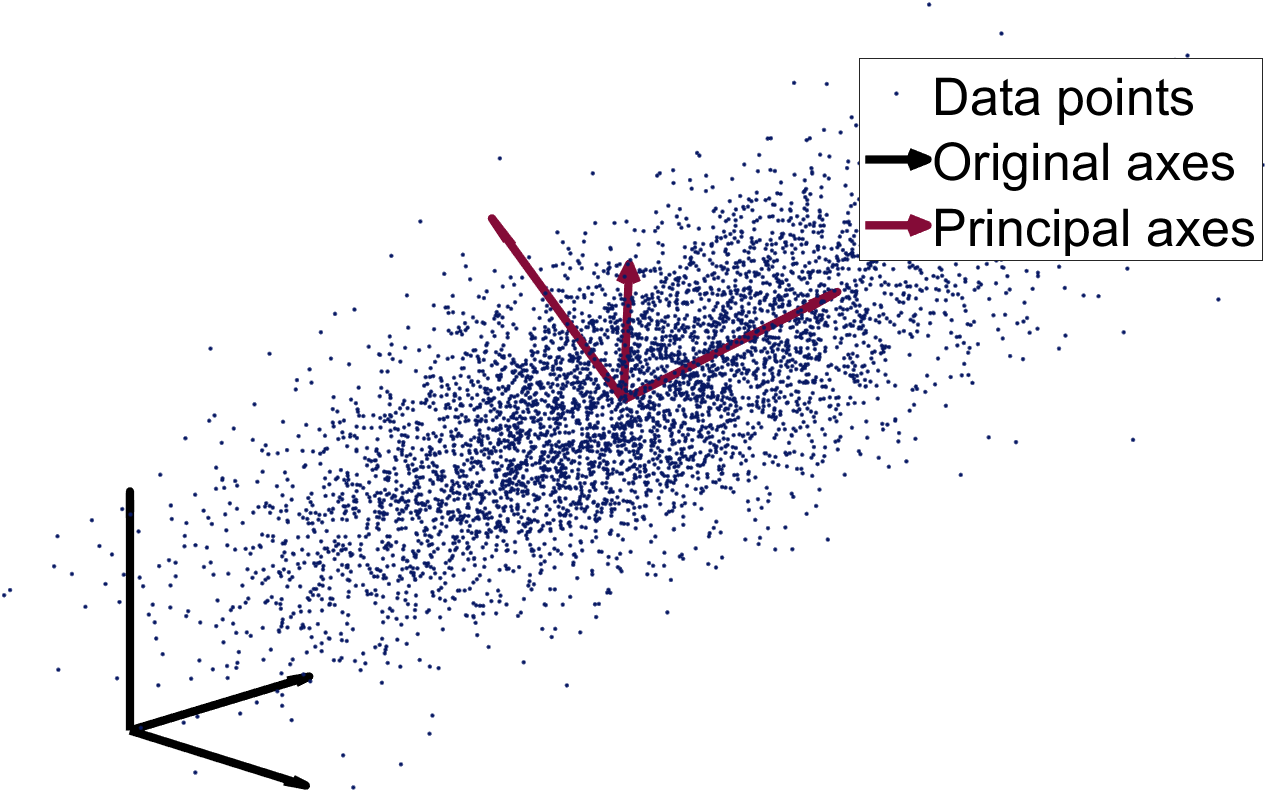}}
    \end{center}
   \caption{set of $\mathbb{R}^3$ scattered points with their original axes ``in black'' and their normalized principal axes``in red'' }
\label{fig:cluster}
\end{figure}

\noindent
Benefiting from the time being represented as a monotonically increasing function, the generated events in a small temporal shift will form planar surfaces (as the surface of extruded material). 
Based on this property, Benosman et al \cite{benosman13}, propose a local plane fitting method where a mapping of clustered events in a spatio-temporal neighborhood can be used to fit a plane (in $x$, $y$ and $t$ coordinates system). The $x$ and $y$ components of the vector normal to the plane represent an estimate of the optical flow. 
\noindent
To make sure the estimated plane is the correct fit, an iterative scheme is used to reject events that are very far from the plane estimate until convergence is met. 
Local Plane fitting methods provide acceptable optical flow accuracy while being intuitively more realistic to describe events' nature, although tuning for $\Delta t$ is also required and plane estimation could be deteriorated due to falsely generated events. 
To alleviate the problem of $\Delta t$ fine-tuning, Mueggler et al \cite{mueggler}, use the so-called active events surface (a surface that stores only the last generated events at each pixel) to cluster events in the neighborhood and then adopt a RANSAC scheme for a better plane fitting. 
In \cite{mueggler}, to refine the output and get a better event's lifetime estimate (which is also the main purpose of our work) an optimization scheme is proposed. 
This scheme is able to smooth the total output of optical flow but did not provide higher accuracy while increasing more computation time.\\

\noindent
Many other contributions in the field of event-based optical flow estimation have been proposed \cite{bardow2016, barranco2015bio, stoffregen2018simultaneous}. 
These methods differ from one another in the use of the type of domain (spatial or frequency) or the segmentation method (based on contrast maximization optimization or spatio-temporal regularization). All these existing methods provide good accuracy but suffer from extremely heavy computation time. 
Since it is accepted that these algorithms will be integrated in more complex schemes to achieve the given tasks, they are considered out of interest for real-time robotics applications even if accuracy and robustness are guaranteed.\\

\noindent
In this paper, we present an optical flow algorithm that can provide accurate, yet simple to implement, estimates while maintaining relatively low computational time. 
The paper is organized as follows: in section \ref{section:nature}, we present the specific properties of event-based cameras and the reason why they are well suited to provide good optical flow algorithm performance. 
Section \ref{section:PCA} demonstrates the concept of Principle Component Analysis (PCA) and the benefits from its adaptation to event-based optical flow scheme. 
Results are shown in section \ref{section:results}. Section \ref{section:conclusion} summarizes the work presented in this paper and the proposed future work.
\begin{figure}[t]
    \begin{center}
        \fbox{\includegraphics[width=0.9\linewidth]{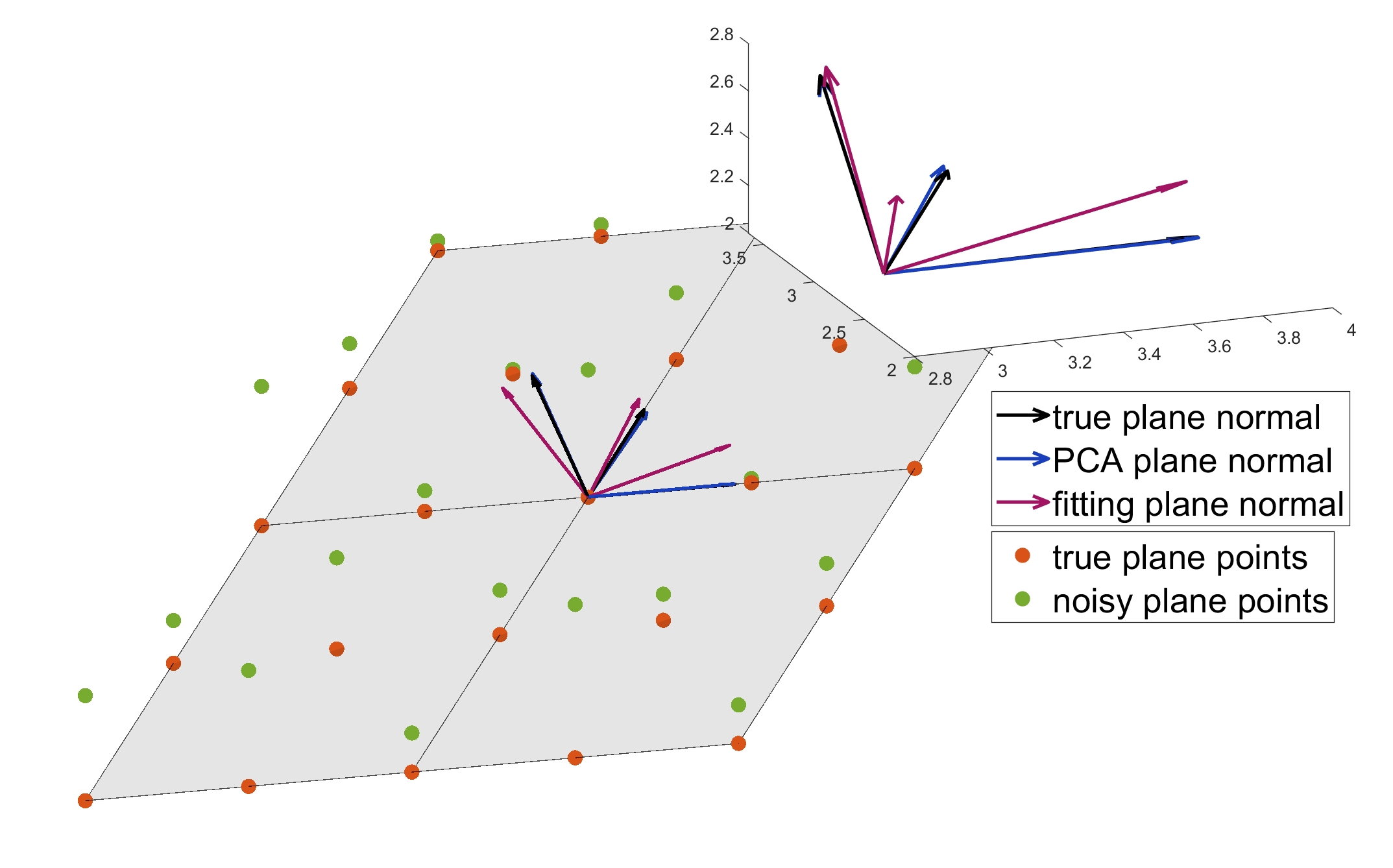}}
    \end{center}
   \caption{In red: points spanning the real actual plane. In green: noisy points representing a plane consensus. A closer look of all the coordinates systems shows up to the right where, aside from plane fitting axes, PCA and the actual plane coordinates look almost identical. In black: the actual plane coordinates system. In blue, the estimated coordinates system estimated using PCA. In red: the coordinates system estimated using plane fitting least square. 
   }
\label{fig:plane}
\end{figure}
\section{Event-Based Nature}
\label{section:nature}
Neuromorphic vision sensors, also called event-based cameras, feature pixels that operate asynchronously and independently. 
Each pixel responds to log intensity change in the environment and triggers events whenever a threshold is attained. 
The asynchronous and independent nature of event-based cameras privileges them thanks to their very high temporal resolution where many events may be triggered in a microsecond. 
Each event is encoded as a tuple $\langle x,y,t,p\rangle$ where $x$ and $y$ are the pixel coordinates, $t$ is the timestamp of the triggered event and $p$ is the polarity in $\{1,-1\}$ that corresponds to positive or negative luminosity change respectively. 
An event is triggered whenever a change exceeds a certain threshold $\delta_t$ which may vary between $10$ and $15\%$ of the luminosity of the last triggered event at each pixel according to the following equation:
\begin{equation}
    \Delta L(x_i,y_i,t_i) = L(x_i,y_i,t_i) - L(x_i,y_i,t_i-\Delta t) = p_i\delta_l
\end{equation}
where $L(x_i,y_i,t_i)$ is the log intensity at the current time and $L(x_i,y_i,t_i-\Delta t)$ is the log intensity of the event created previously at this pixel at $t_i - \Delta t$. 
Changes in luminosity are supposed to be only due to movements since small time intervals are considered. 
Events encompass optical flow information as it can be shown by approximating $\Delta L(x_i,y_i,t_i)$ using Taylor expansion:
\begin{equation}
    \Delta L(x_i,y_i,t_i) \approx \frac{\partial L}{\partial t}(x_i,y_i,t_i)\Delta t = \frac{\partial L}{\partial x}\frac{\partial x}{\partial t} + \frac{\partial L}{\partial y}\frac{\partial y}{\partial t} + \frac{\partial L}{\partial t} \\
\end{equation}
Considering that event-based cameras encapsulate optical flow information and provide quasi-continuous signals, suggests pushing the limits of classical optical flow algorithms by proposing new adaptations to event-based cameras. Even though many improvements for event-based optical flow have been introduced in the state-of-the-art, many restrictions like accuracy and computational power still prevent the adoption of optical flow with neuromorphic vision sensors. As a step to overcome these burdens,  a novel algorithm for optical flow estimation is presented in the next section.

\section{PCA Optical Flow}
\label{section:PCA}
PCA \cite{pearson1901} was first introduced to the scientific community by Karl Pearson as a linear dimensionality reduction method. The principle is to map higher dimensional spaces $\mathbb{R}^n$ data to lower ones by providing a hierarchical orthogonal coordinates system, changing the basis that spans data dimensions\footnote{PCA provides hierarchical principal axes where each axis is scaled to span a principal dimension of the data.} (see Figure \ref{fig:cluster}). Dimensionality reduction is done using PCA by maximizing the variance of the data around the principal axes in the lower dimension space. 
Since variance is maximized around the singular vectors representing the data, eigenvalue decomposition can be used to find the principal components of the data.\\

\noindent
Based on the hypothesis that triggered events create a plane in a small neighborhood, PCA can be employed to find the two orthogonal vectors spanning the created plane and the third will be the plane's normal\footnote{plane normal corresponds to 0 variance for a perfect plane (no thickness for flat surfaces).}. 
PCA is found to be able to estimate the plane's normal and provide better accuracy than least-square plane fitting algorithms provided in the state of the art \cite{benosman13, mueggler, rueckauer} (see Figure \ref{fig:plane}). 
Applying PCA in an optical flow event-based scheme is carried out in three steps. First, events provided by the camera are conditioned to put aside events that are redundant or created by noise or strong luminosity changes. Second, the optical flow is estimated using PCA method on conditioned events. Finally, the optical flow is regularized to assure robustness of the estimated optical flow. These steps are detailed extensively in the following section.\\
\subsection{Events Filtering}
Provided events on pixel-level are due to changes in the environment makes event-based cameras highly prone to different noise sources.
Events filtering step is required before applying any algorithm to verify that triggered events correspond to real changes in the environment. 
First, positive and negative events are separated and processed independently where a refractory filter filters out events that may be triggered due to sharp changes in intensity. 
Events are filtered if an event has already been triggered on the same pixel within a certain time interval. 
The limit is chosen to be 20 $ms$ for events triggered of the same polarity and 1 $ms$ for opposite polarity. These values are chosen based on practical experiments. To make sure that all events belong to real motion, an adaptive activity filter \cite{khairallah2021} is applied. 
Adaptive filtering in this context consists of storing 8-pixel neighborhood events in a buffer and checking whenever a new event is triggered. For each event, the time difference between it and at least 3 of the 8-pixel neighborhood events in the buffer does not exceed the adaptive support time $T_f$ which is computed as a function of events frequency\footnote{Events frequency varies according to the environment dynamics which affects event}:
 \begin{eqnarray}
       \alpha & = & \frac{k}{\log f_e}\\
       T_f & = & \frac{T_{max} - T_{min}}{\alpha_{max} - \alpha_{min}} (\alpha - \alpha_{min}) + T_{min}
       \label{eq:1}
 \end{eqnarray}

\begin{algorithm}[t]
    \SetAlgoLined
    \KwData{$e\{x,y,t,p\}_{i=0}^{N-1}$}
    \KwResult{$e\{x,y,t,p,v_x,v_y\}_{i=0}^{N-1}$}
    \textbf{Initialize:} $\epsilon\;, n = \#\;of\;pixels,\;ev = zeros(col,row,2)\;,$\\ 
    $flow = zeros(col,row,2,3), $\\ 
    $t_s = 20\;ms\;,\;t_o = 1\;ms$\\
    \For{$i\gets0$  \KwTo $N-1$}{
    $flag = 0$\\
    $x,y,t,p = e_i$\\
    $p_o = opposite\;polarity$\\
    \If{$(t_i-ev(x_i,y_i,p_i)<t_s$) \& $(t_i-ev(x_i,y_i,p_o)<t_o)$}{ 
    $T_f$ = adaptive time (eqn(3), (4))\\
     \If{ in neigh $(t_i - t_{neigh})<T_f$}{
                $flag = 1$\\
                $ev(x_i,y_i,p_i) = t_i$
        }
    }
    \If{$flag = 1$}{
    $neigh\gets $ events in $ev$ in $n\times n$ neighborhood\\
    \eIf{$size(neigh)>3$}{
    \textbf{Estimate:} $\mathbf{\mu_x, \mu_t, \mu_t}$\\
    \textbf{Construct:} $N$ = eqn(5), $\Sigma$ = eqn(6)\\
    \textbf{Compute:} [V,E] = eig($\Sigma$)\\
    \eIf{$e(1,1)<\epsilon$}{
    [$v_{xi},v_{yi}$] = eqn(7)\\
    }{$[v_{xi},v_{yi]} = [0,0]$}
    }{$[v_{xi},v_{yi]} = [0,0]$}
    }}
    \caption{PCA Optical Flow Estimation}
    \label{alg:compute PCA}
\end{algorithm}
\begin{figure}[t]
    \begin{center}
        \fbox{\includegraphics[width=0.9\linewidth]{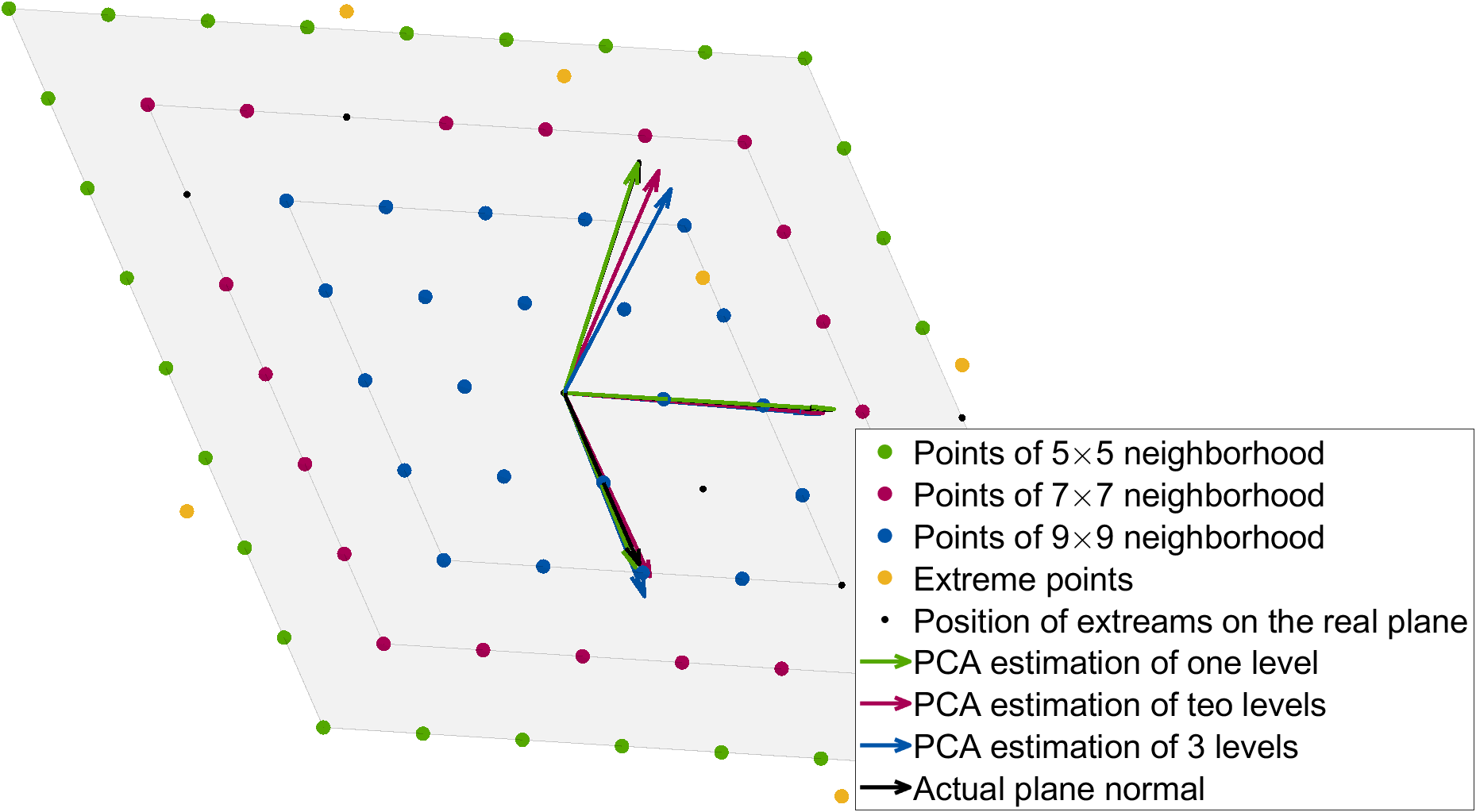}}
    \end{center}
   \caption{Three levels of neighborhood are shown, $5\times 5$ neighborhood "in blue", $7\times 7$ neighborhood "in red", $9\times 9$ neighborhood "in green" and the extreme points (outliers) "in yellow". 
   PCA estimation of the three levels are shown.}
\label{fig:neigh}
\end{figure}

\subsection{PCA Optical Flow Estimation}
PCA was first adapted to event-based nature in a line segmentation scheme \cite{everding2018low}, here we adapt it for optical flow estimation. We estimate best plane fit of triggered events to obtain spatial components where the plane is expressed as:
\begin{equation}
    \begin{bmatrix}
    a\\b\\c\\d
    \end{bmatrix} \begin{bmatrix}
    x & y & t & 1
    \end{bmatrix} = 0
\end{equation}
$(a\;b\;c\;d)$ are the estimated plane parameters. Principal components of a set of data points can be computed by finding the eigen vectors of the covariance matrix $\mathbf{\Sigma}$ of this set of centralized points. 
The first principal component (vector) will be the one corresponding to the largest eigenvalue and the next ones correspond to lower eigenvalues respectively. 
For a set of $n$ polarized events  $\{x_i, y_i, t_i\}$ created in a neighborhood $\mathbf{N}$ in $\mathbb{R}^3$, they are centralized by subtracting the mean of each dimension such that:
\begin{equation}
    \mathbf{N} = \begin{bmatrix}
x_1 - \mathbf{\mu_x} & y_1 - \mathbf{\mu_y} & t_1 - \mathbf{\mu_t}\\
\vdots & \vdots & \vdots\\
x_n - \mathbf{\mu_x} & y_n - \mathbf{\mu_y} & t_n - \mathbf{\mu_t}\\
\end{bmatrix}
\end{equation}
The matrix $\mathbf{N}$ is used to construct the covariance matrix as:
\begin{equation}
    \mathbf{\Sigma} = \begin{bmatrix}
    \mathbf{\sigma_{xx}} & \mathbf{\sigma_{xy}} & \mathbf{\sigma_{xt}} \\
    \mathbf{\sigma_{yx}} & \mathbf{\sigma_{yy}} & \mathbf{\sigma_{yt}} \\
    \mathbf{\sigma_{tx}} & \mathbf{\sigma_{ty}} & \mathbf{\sigma_{tt}} \\
\end{bmatrix}
\end{equation}
where $\mathbf{\sigma_{ij}} = \sum_{k=0}^n i_kj_k$. 
The eigen vectors of $\mathbf{\Sigma}$ represent the principal components spanning the neighborhood, two orthogonal vectors span the plane and the third one corresponding to the smallest eigenvalue is perpendicular to the plane. 
The vector $\mathbf{V_p}$ corresponding to the smallest to eigenvalue is considered the plane normal. This vector can be used directly to estimate the optical flow, however we employ a step to validate the accuracy of estimated parameters first. We estimate the plane parameters as follows;
\begin{eqnarray}
      &a = V_{px}\;,\;b = V_{py}\;,\;c = V_{pt}\\
      &d = -(V_{px}x + V_{py}y + V_{pt}t)
\end{eqnarray}
$(V_{px},V_{py},V_{pt})$ are the components of the vector $V_p$, $(x,y,t)$ are the coordinates of the event under test. The plane parameters are used to evaluate $t_{est}$ value for each pixel used to fit the plane where only the event spatial coordinates are used.
\begin{equation}
    t_{est} = -\frac{ax + by+ d}{c}
\end{equation}
if the absolute difference between $t_{est}$ and the actual triggering time of the event $t$ exceeds a threshold $\delta$ the event is considered outlier. A consensus of inliers should be more than $(1-\epsilon)N^2/2$ where $\epsilon$ is the accepted ratio of outliers, otherwise the estimated plane will be rejected. The estimation of optical flow is obtained as follows:
\begin{equation}
    \begin{bmatrix}
    v_x\\v_y
    \end{bmatrix} = \frac{-V_{pt}}{V_{px}^2 + V_{py}^2 } \begin{bmatrix}
    V_{px}\\V_{py}
    \end{bmatrix}
\end{equation}
Hence the lifetime of each event is obtained:
\begin{equation}
    t_{lifetime} = \frac{V_{px}^2 + V_{py}^2}{V_{pt}}
\end{equation}
Algorithm \ref{alg:compute PCA} demonstrates a step-by-step pseudo-algorithm of our method.\\

\subsection{Spatio-Temporal Optical Flow Regularization}
Due to the sparsity and noisy nature of neuromorphic sensors, it is required to go the extra mile and refine the estimated optical flow to make sure it depicts the true optical flow. 
We try two different spatio-temporal regularization techniques to see which would work best w.r.t estimation quality and computation time.\\
\begin{figure}[t]
\begin{center}
    \fbox{\includegraphics[width=0.9\linewidth]{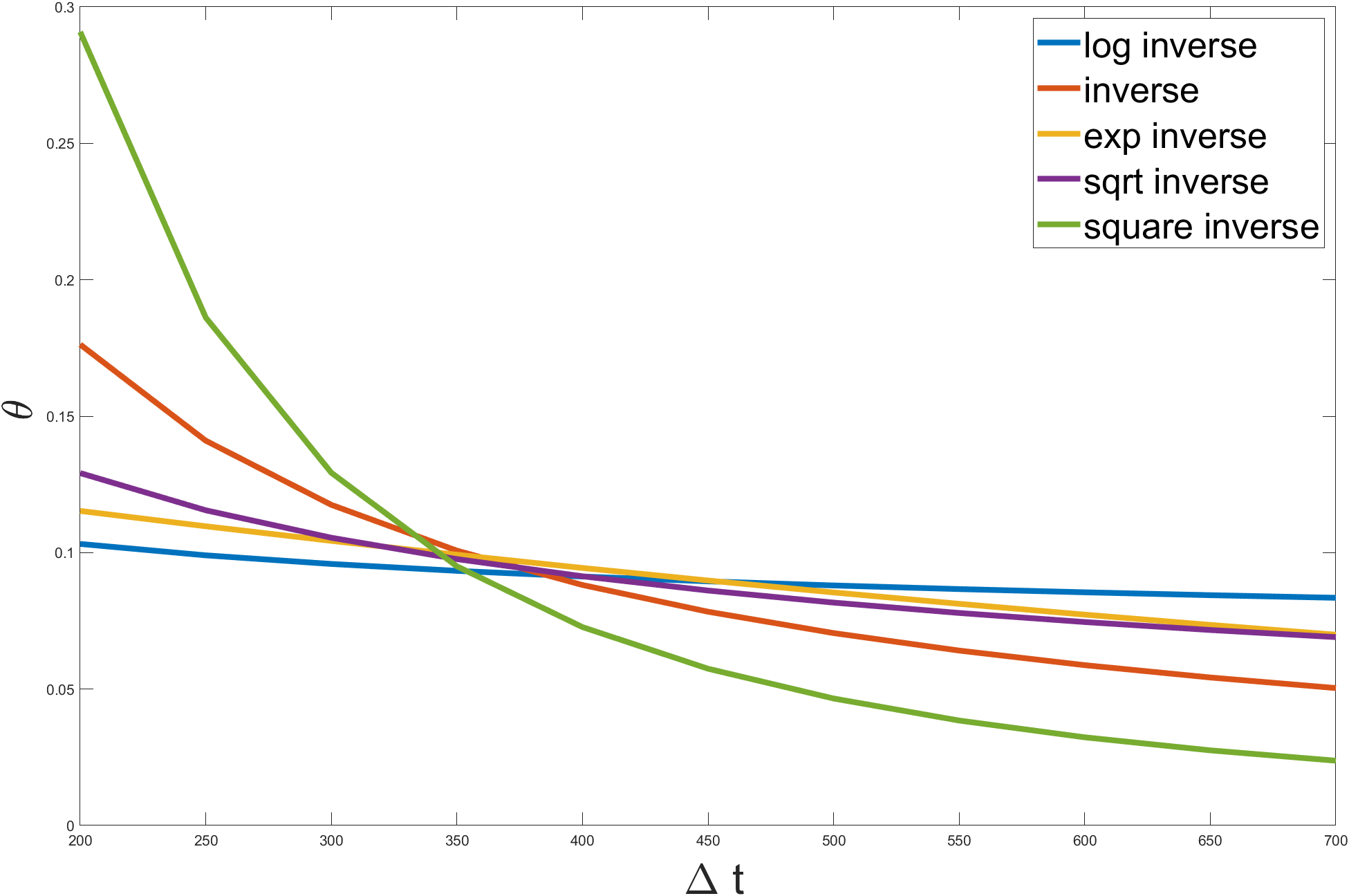}}
\end{center}
\caption{Variation of weighting functions plotted according to real data collected. Inverse exponential (in green) gives extreme weights, inverse logarithmic gives averaging-like effect and the inverse gives weightings to be distributed in between.}
\label{fig:weightings}
\end{figure}
\noindent
Methods in the state-of-the-art demand solving a system of equations that depends on the size of the chosen neighborhood. The first technique benefits from the fact that the neighborhood's size will not affect much the computation time for PCA\footnote{we always compute eigen vectors of a $3\times 3$ matrix which would take most of the time}.  We use different levels of the neighborhood to estimate the optical flow and take the mean as the best estimate of optical flow to be sure that extremes are filtered out (see Figure \ref{fig:neigh}).\\

\noindent
The other technique consists of storing the optical flow in a buffer called ``active optical flow'' and then weightings are given to the optical flow estimated previously in a specific neighborhood of the event under test.
Weightings $\theta_i$ are chosen to be inversely proportional to the difference of timestamps of the event under test and events triggered in the neighborhood so that the final optical flow estimation will be:
\begin{equation}
\begin{bmatrix}
v_x\\v_y
\end{bmatrix} = \sum_{i=1}^n \theta_i \begin{bmatrix}
v_{xi}\\v_{yi}
\end{bmatrix}
\end{equation}
\noindent
Allegedly, choosing the inverse of exponential would be a reasonable choice for its vanishing values for higher time difference. The choice of this function is found to greatly suppress the effect of old optical flow on the neighborhood (see Figure \ref{fig:weightings}).
Among all possible weighting functions, we chose to use
the inverse of the difference  because it gives relatively correctly distributed weightings while being the fastest to compute. It has been found that choosing a smaller neighborhood for weightings\footnote{i.e. $5\times 5$ for a $7\times 7$ neighborhood to estimate optical flow.} than the neighborhood chosen to estimate the optical flow gives better results. 
The regularization process is shown in Algorithm \ref{alg:regularize}.\\

\begin{algorithm}[bt]
    \SetAlgoLined
    \KwData{$[v_x,v_y,t],\;flow$}
    \KwResult{regularized $[v_x,v_y]$}
    \textbf{Initialize:} $reg$,   $n_l = \#\;of\;levels$\\
   \eIf{$reg = weighting$}{
    $t_n \gets$ times in $n\times n$ neighborhood in $flow$
    $v \gets [v_x,v_y]$ in $n\times n$ neighborhood in $flow$\\
    $W = \frac{1}{t-t_n}$\\
    $W = \frac{W}{norm(W)}$\\
    $[v_x,v_y]_{reg}$ calculated using eqn(8)}{
    $v \gets [v_x,v_y]$
    \For{$i\gets 1$  \KwTo $n_l$}{
    \textbf{Recall:} Algorithm(\ref{alg:compute PCA}) with $n = n-1$\\
    $v \gets v + [v_x,v_y]$\\
    $[v_x,v_y]_{reg} = \frac{v}{n_l}$}
    }
    
    \caption{PCA Optical Flow Regularization }
    \label{alg:regularize}
\end{algorithm}
\noindent
We tested the availability to apply PCA optical flow algorithms with and without those different regularization techniques and compare them with state-of-the-art event-based optical flow algorithms and show the results in the following section.

\section{Results}
\label{section:results}
To validate the reliability and accuracy of the algorithm we propose, we used the data-sets provided previously \cite{khairallah2021, mueggler}. 
We used two sequences of the first data-set which cover translational and rotational dominant motion and the data set of moving stripes to validate the estimated lifetime. In \cite{khairallah2021}, a VICON system is used to track the motion of multiple objects and 3D/2D projection is used to create optical flow ground truth (see Figure 6).
These sequences are recorded using an ATIS (Asynchronous Time-based Image Sensor) camera \cite{atis} provided by PROPHESEE with a $480\times 360$ resolution which can provide change detection, events detection and exposure measurements (see Figure \ref{fig:davis}). 
The camera is handheld and tracks a checkerboard while the desired motion is fulfilled. 
We test the validity of PCA optical flow with and without regularization and compare it with the methods of event-based Lucas-Kanade and plane fitting methods adapted with Savitzky-Golay filter as described in \cite{rueckauer2016}. Usage of Savitzky-Golay filter is proven to improve accuracy and reduce computation time.
Experiments are carried out with a spatial neighborhood of $7\times 7$, three levels for pyramidal regularization are chosen and $\Delta t$ for Lucas-Kanade method is chosen to be $10\;ms$.\\ 

\noindent
Four metrics are used to provide a quantitative and comprehensive comparison. The Average End Point Error (AEPE) measures differences in optical flow magnitude, The Average Angular Error (AAE) measures differences in optical flow orientation, Lifetime error to check the difference between estimated and real lifetime of each event and computational time shows the availability to use these algorithms in real-time applications. Table \ref{tab:computationtimes} summarizes the computation time required to estimate optical flow for each algorithm. Table \ref{tab:comp1} shows AEPE and AAE performance metrics evaluated using our recorded sequences. 
\begin{figure}[t]
    \begin{center}
        \fbox{\includegraphics[width=0.9\linewidth]{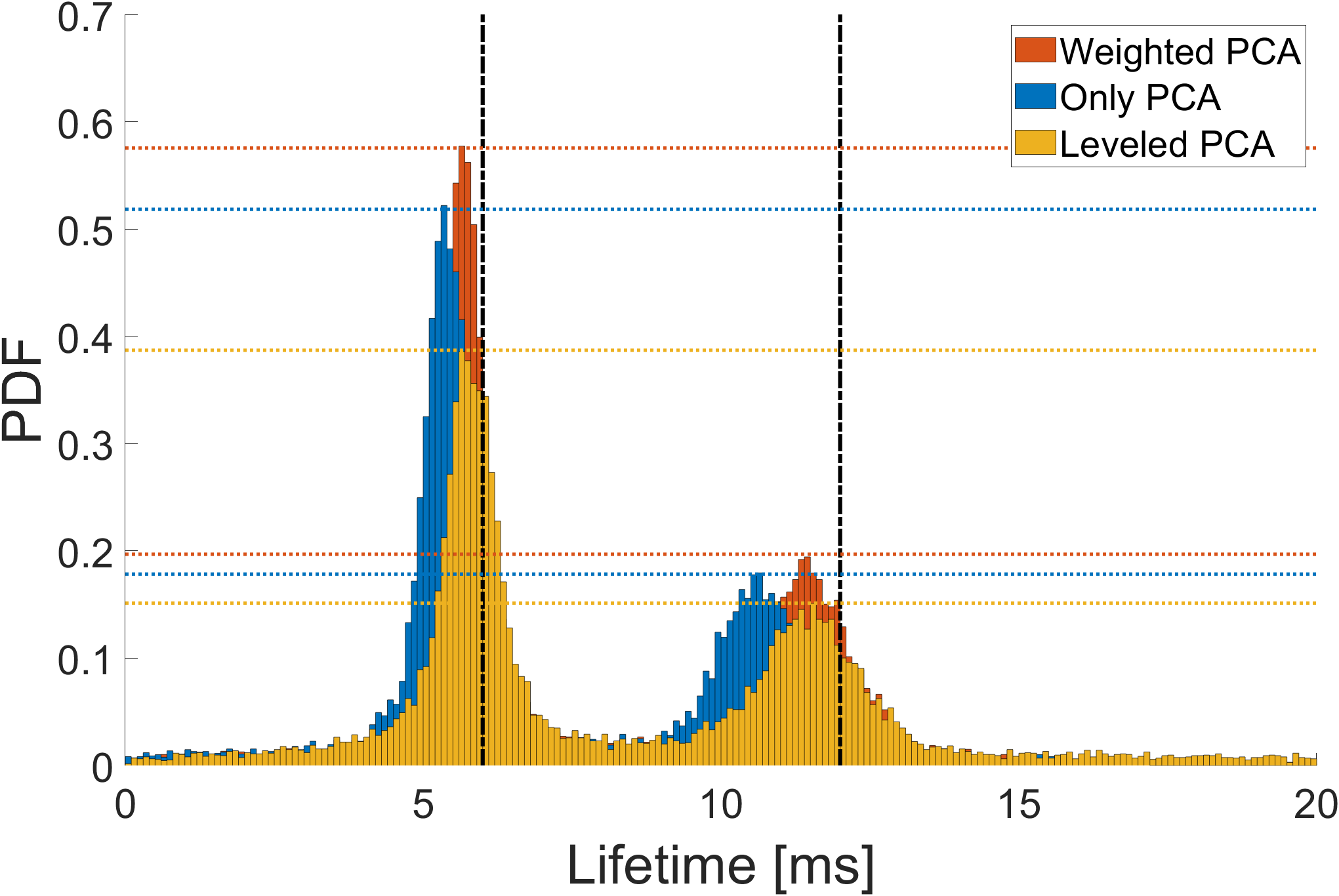}}
    \end{center}
   \caption{Probability Density Function(PDF) of the estimated lifetime for only PCA (Blue), weighted PCA (Red) and leveled PCA (Yellow) of the sequence \cite{mueggler} \tt{stripes}}
\label{fig:hist}
\end{figure}
\begin{table}[t]
    \centering
    \small
    \begin{tabular}{|c|c|c|c|c|}
    \hline
    \multirow{3}{*}{Algorithm} & \multicolumn{2}{c}{$\tt{slow\_stripes}$}\vline & \multicolumn{2}{c}{$\tt{fast\_stripes}$}\vline \\
    \cline{2-5}
     & max bin  & \multirow{2}{*}{\%} & max bin & \multirow{2}{*}{\%} \\
     & (ms) & & (ms) & \\
    \hline
    PCA Only & 5.35  & 52.19 & 10.65 &  17.95 \\
    Weighted PCA & 5.55  &57.72& 11.05 & 19.44 \\
    Leveled PCA & 5.65  &38.58& 11.45 & 15.11 \\
    \hline
    \end{tabular}
    \caption{The maximum bin value for the life time estimation of the dataset provided in \cite{mueggler} and the percentage of each bin}
    \label{tab:my_label11}
\end{table}
\subsection{Average End Point Error}
The average end point error is defined as: 
\begin{equation}
    AEPE = \frac{1}{N} \sum \limits_{i=1}^N ||\textbf{u}_i -\hat{\textbf{u}}_i ||
\end{equation}
where $\textbf{u}_i$ and $\hat{\textbf{u}}_i$ are the estimated and the ground truth optical flow respectively. 
AEPE illustrates the error in the magnitude of estimated optical flow.\\
\begin{figure*}[htb]
    \label{fig:results1}
    \centering
    \fbox{
        \parbox{1\textwidth}{\centering
        \includegraphics[width=0.33\linewidth]{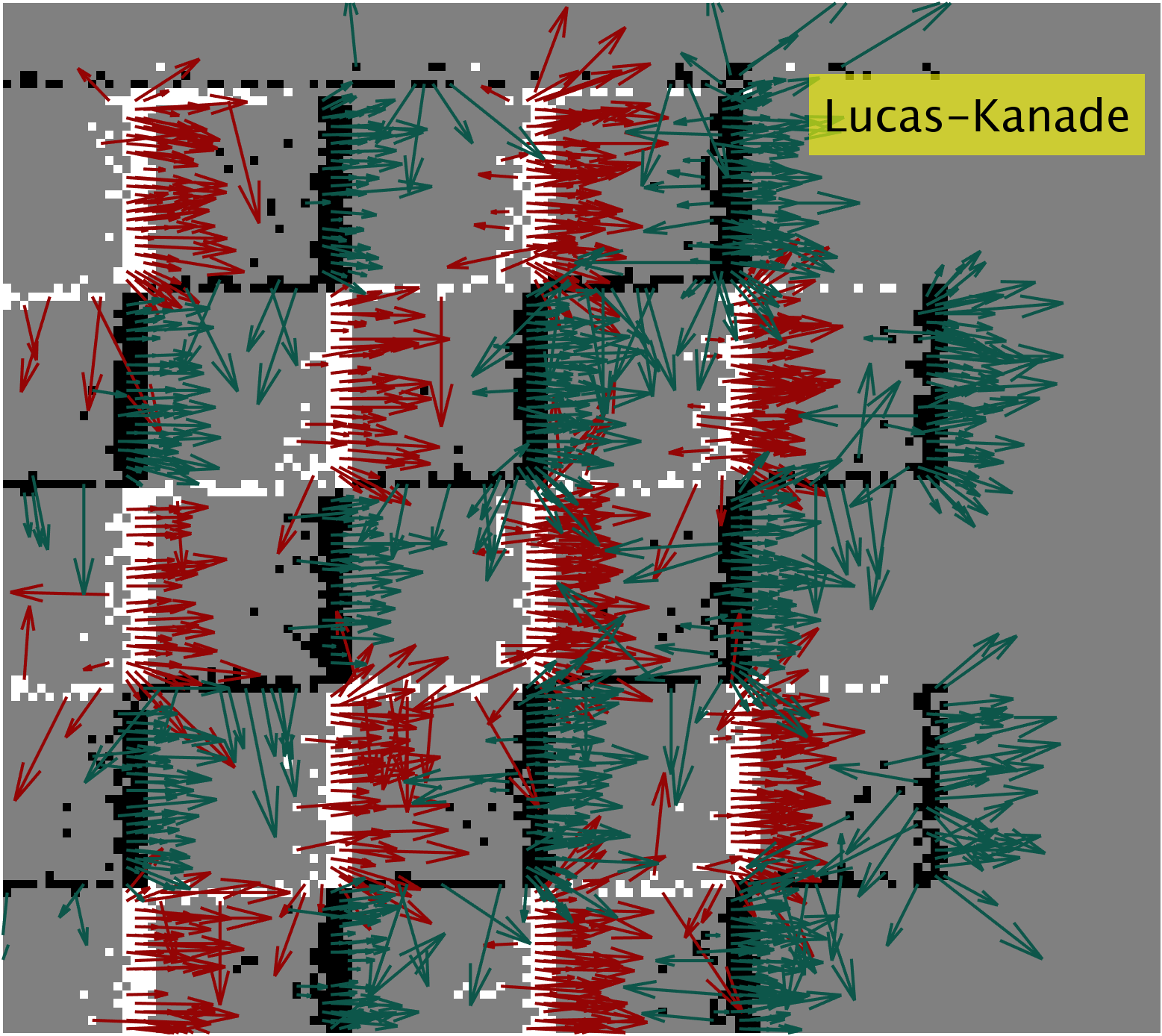} 
        \includegraphics[width=0.33\linewidth]{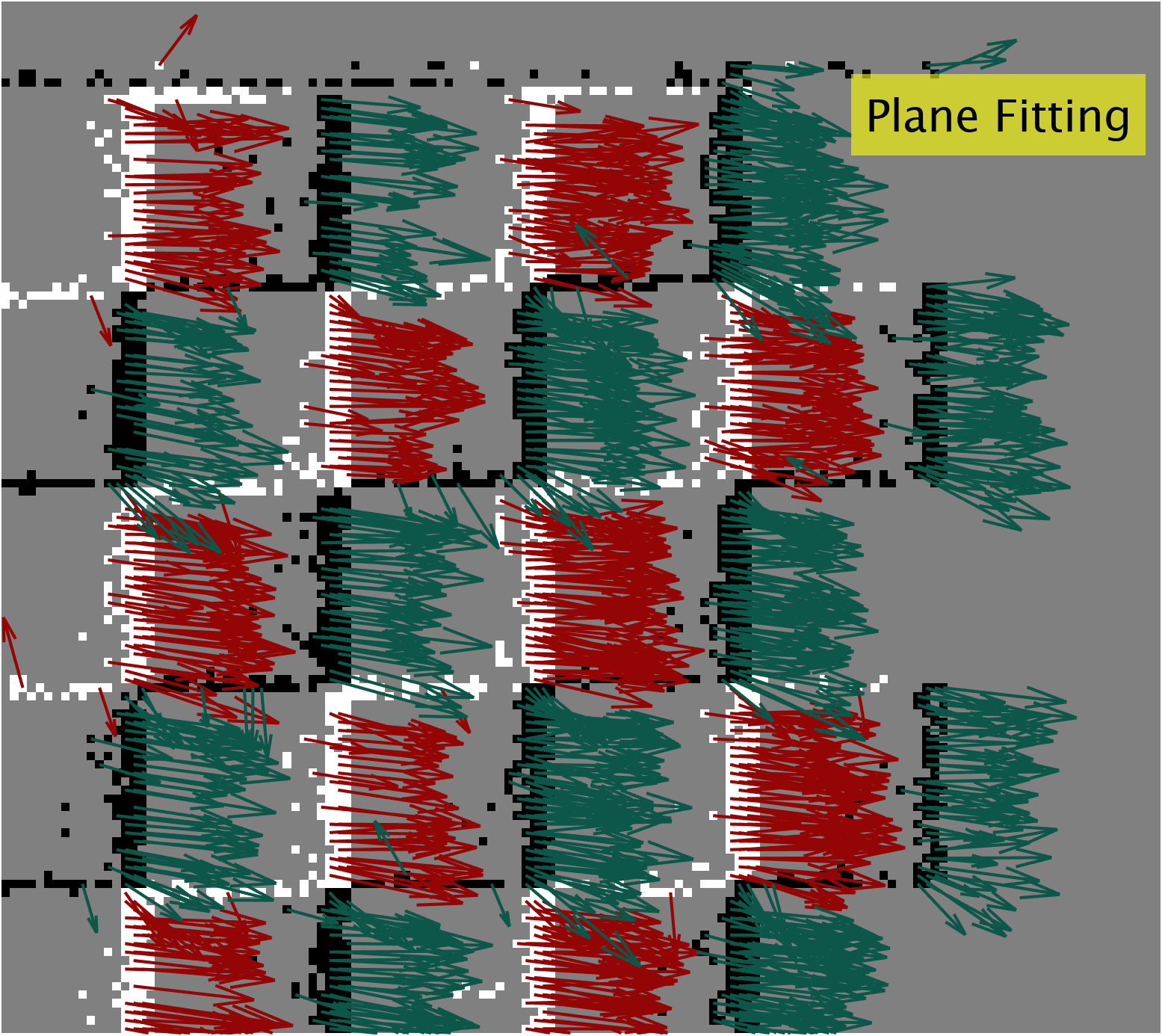}
        \includegraphics[width=0.33\linewidth]{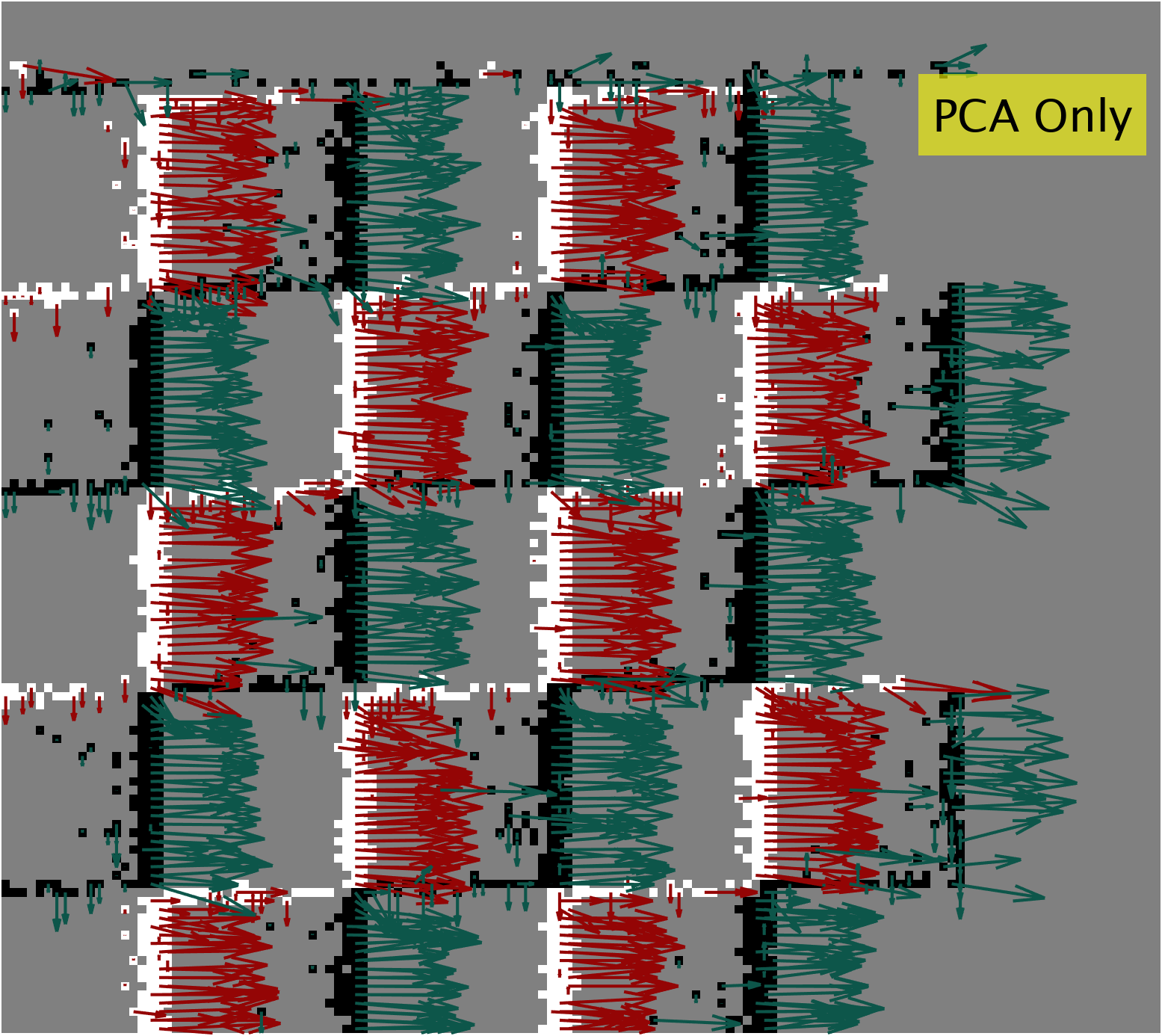}
        \includegraphics[width=0.33\linewidth]{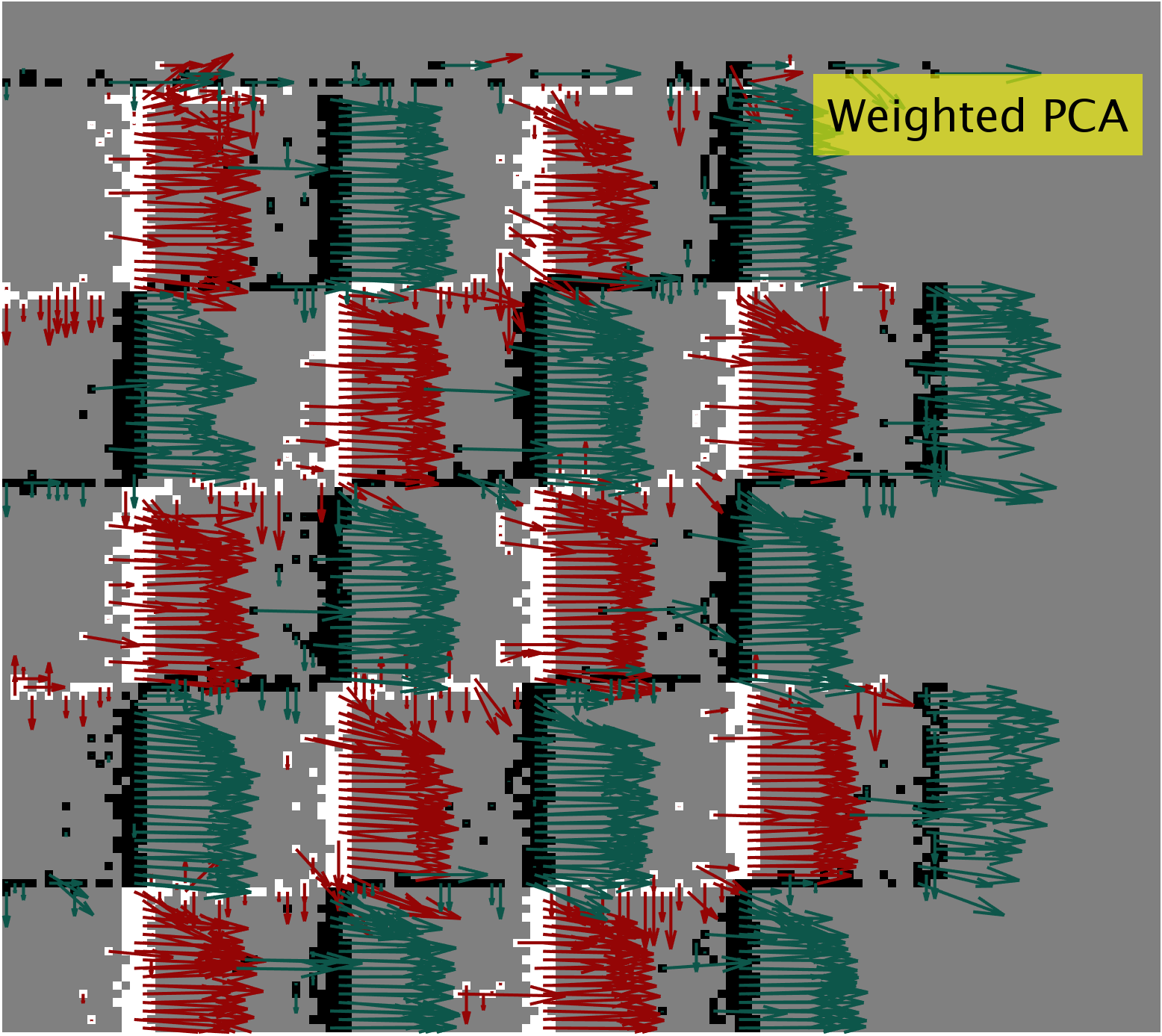}
        \includegraphics[width=0.33\linewidth]{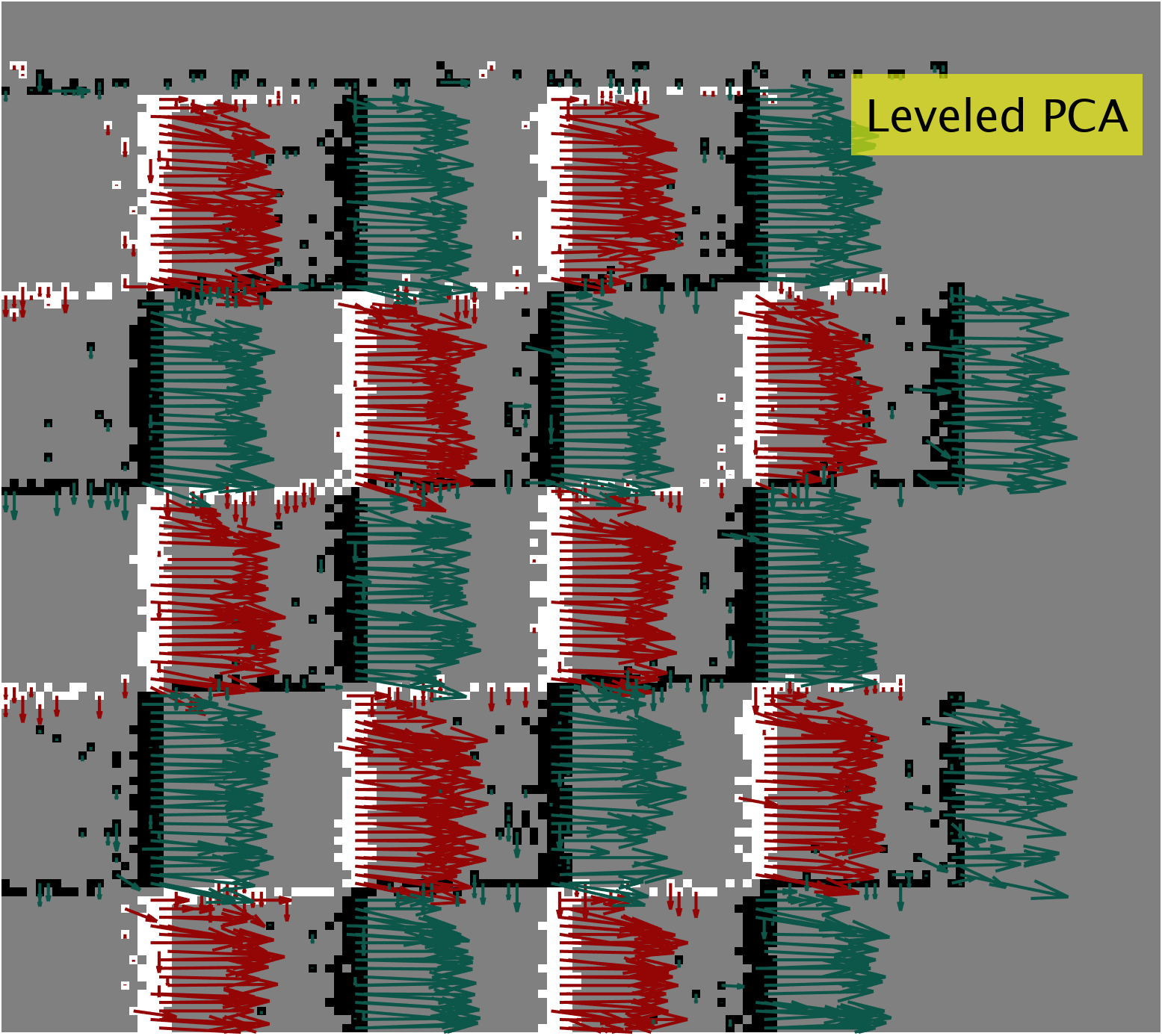}
        \includegraphics[width=0.33\linewidth]{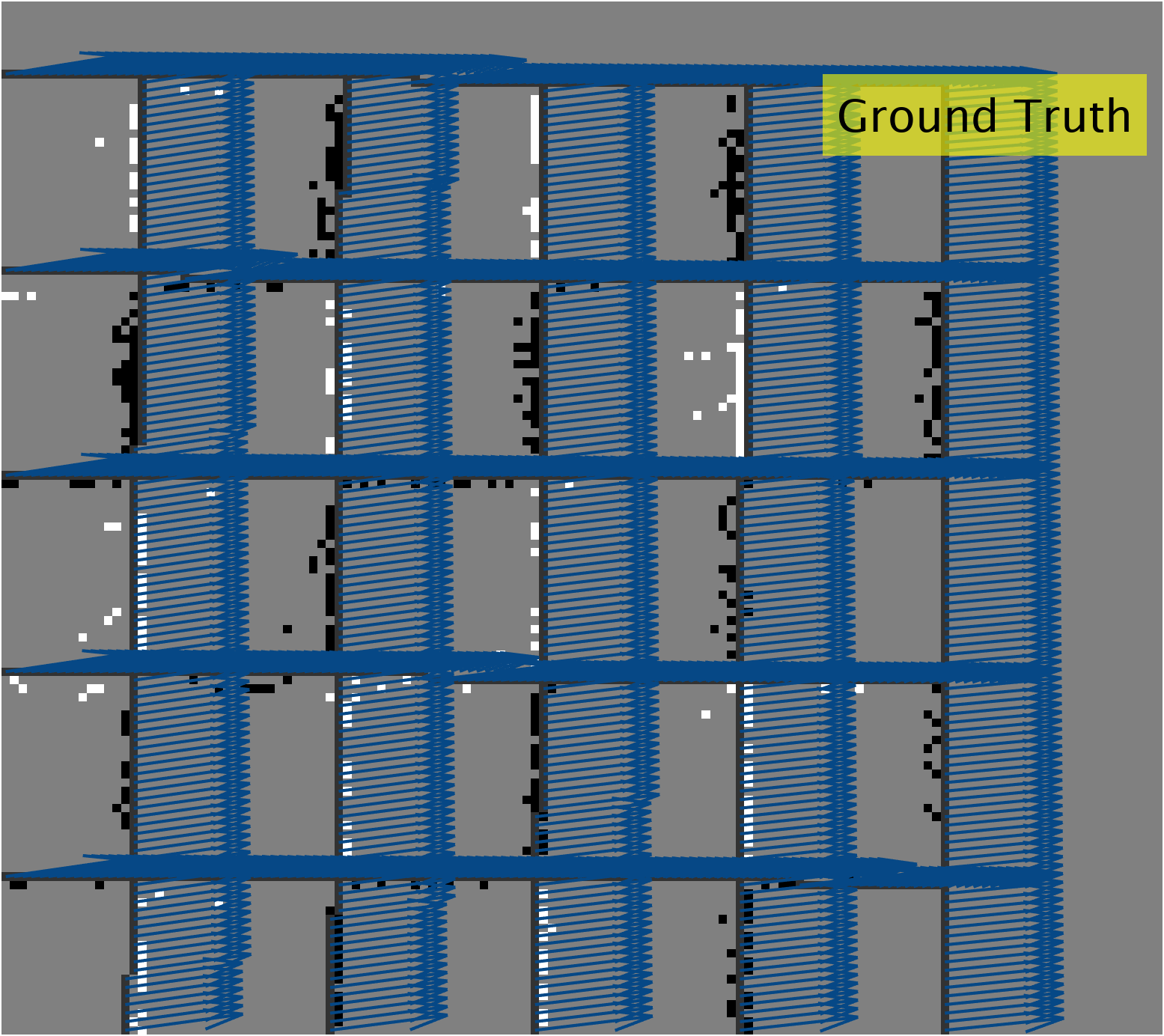}
        }
    }
    \caption{the results obtained from our algorithm and methods in the state of the art, up from left to right: lucas-kanade with  Savitzky Golay filter, plane fitting with Savitzky Golay filter and PCA with no regularization. down from left to right, PCA with weighting regularization, PCA with pyramidal levels regularization and finally the synthitically generated ground truth is displayed on top of real events. Red arrows represent optical flow estimated from positive events, green arrows represent optical flow estimated from negative events. The ground truth is represented by blue arrows.}
\end{figure*}
\noindent
\textbf{Translation Scenario:} The usage of PCA, in general, shows better accuracy in magnitude estimation. 
Applying PCA only improved the AEPE by more than $8\%$ w.r.t true optical flow compared to the local plane fitting method. 
Applying leveled PCA  regularization provides the best magnitude estimation and improves the overall estimation by more than $11\%$. 
Weighted PCA shows a slight improvement in magnitude estimation compared to PCA only.\\
\begin{table}[htb]
\centering
\begin{tabular}{|c|c|c|}
    \hline
    \multirow{2}{*}{Algorithm} &  \multicolumn{2}{c|}{\tt{translation}} \\
    \cline{2-3}
     & Magnitude (rel.) & Orientation ($\si{\degree}$) \\
    \hline
    PCA only & $0.069 \pm 0.024$ & $7.872\pm4.665$ \\
    PCA /w weights &  $0.061 \pm 0.023$ & $\mathbf{5.671\pm 4.667}$ \\
    PCA /w levels & $\mathbf{0.046 \pm 0.015}$ & $6.599 \pm 5.108 $ \\
    \hline
    Local Plane & $0.158\pm 0.099$ & $13.158\pm 7.542$ \\
    \hline
    Lucas-Kanade & $0.237\pm 0.085$ & $19.540\pm 8.25$ \\
    \hline
    \hline
    \multirow{2}{*}{Algorithm} & \multicolumn{2}{c|}{\tt{rotation}} \\
    \cline{2-3}
     & Magnitude (rel.) & Orientation ($\si{\degree}$) \\
    \hline
    PCA only & $0.081 \pm 0.032$ & $11.854 \pm 4.513$ \\
    PCA /w weights & $0.075 \pm 0.0401$ & $\mathbf{11.236 \pm 3.345}$ \\
    PCA /w levels & $\mathbf{0.071 \pm 0.048}$ & $12.014 \pm 4.868$ \\
    \hline
    Local Plane & $0.173 \pm 0.054$ & $15.568 \pm 6.540$ \\
    \hline
    Lucas-Kanade & $0.275 \pm 0.082$ & $22.634 \pm 8.215$ \\
    \hline
    \end{tabular}
    \caption{Relative average end point error and average angular error for rotational and translational sequences.}
    \label{tab:comp1}
\end{table}

\begin{table}[bt]
    \centering
    \begin{tabular}{|c|c|}
        \hline
        \multirow{2}{*}{Algorithm} &  Computational Time\\
         & per Event ($\mu s$)\\
        \hline
        Lucas-Kanade& $3.32 \pm 1.07$ \\
        \hline
        Plane Fitting & $1.18 \pm 0.74$ \\
        \hline
        PCA only & $\mathbf{0.29 \pm 0.05}$\\
        \hline
        PCA with weights& $0.51 \pm 0.06$ \\
        \hline
        PCA with Levels& $0.78 \pm 0.07$ \\
        \hline
    \end{tabular}
    \caption{Computation times needed for calculations per event.}
    \label{tab:computationtimes}
\end{table}
\noindent
\textbf{Rotational Scenario:} Estimating optical flow in rotation motion is considered more critical because magnitude would vary in a small neighborhood. For this reason, AEPE is shown to be slightly higher than in the translational scenario. Leveled PCA regularization is shown to give the best results and relatively provide the same percentage of improvement in rotational scenarios compared to the state-of-the-art.  

\subsection{Average Angular Error}
The metric to measure the differences of orientation is the average angular error and is defined as:
\begin{equation}
    AAE = \frac{1}{N} \sum\limits_{i=1}^N cos^{-1} \left(\frac{\hat{\textbf{u}}_i^T \textbf{u}_i}{||\hat{\textbf{u}}_i||||\textbf{u}_i||}\right)
\end{equation}
\begin{figure*}[tb]
    \label{fig:lifetimes}
    \centering
    \fbox{
        \centering
        \includegraphics[width=0.25\linewidth]{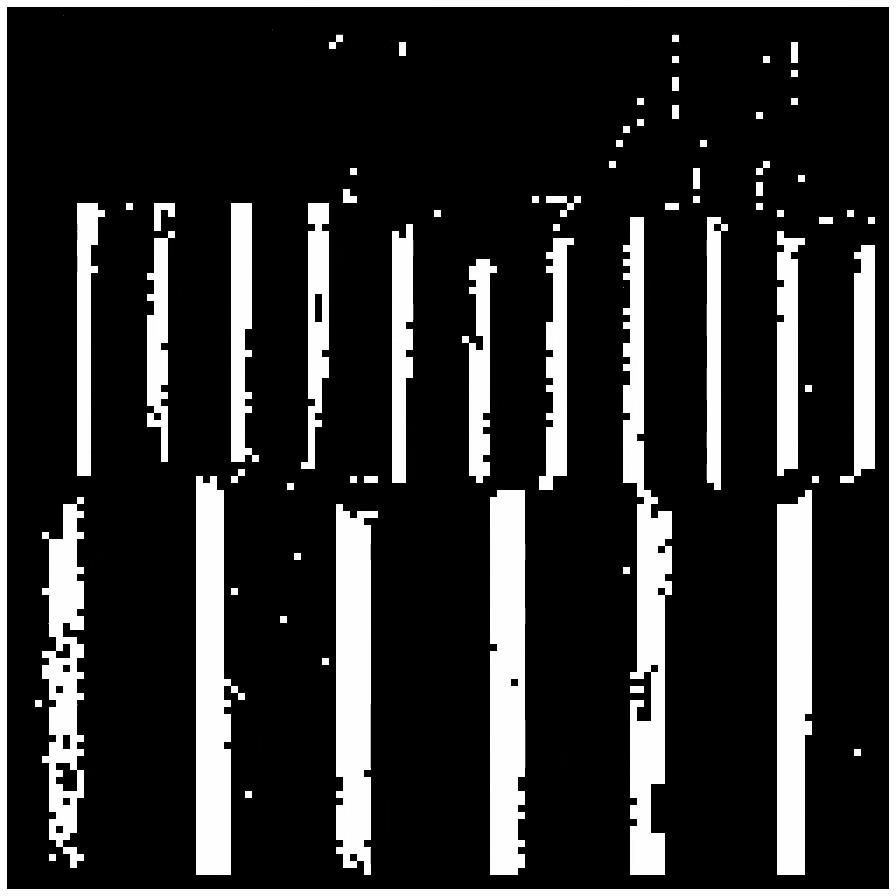} 
        \includegraphics[width=0.25\linewidth]{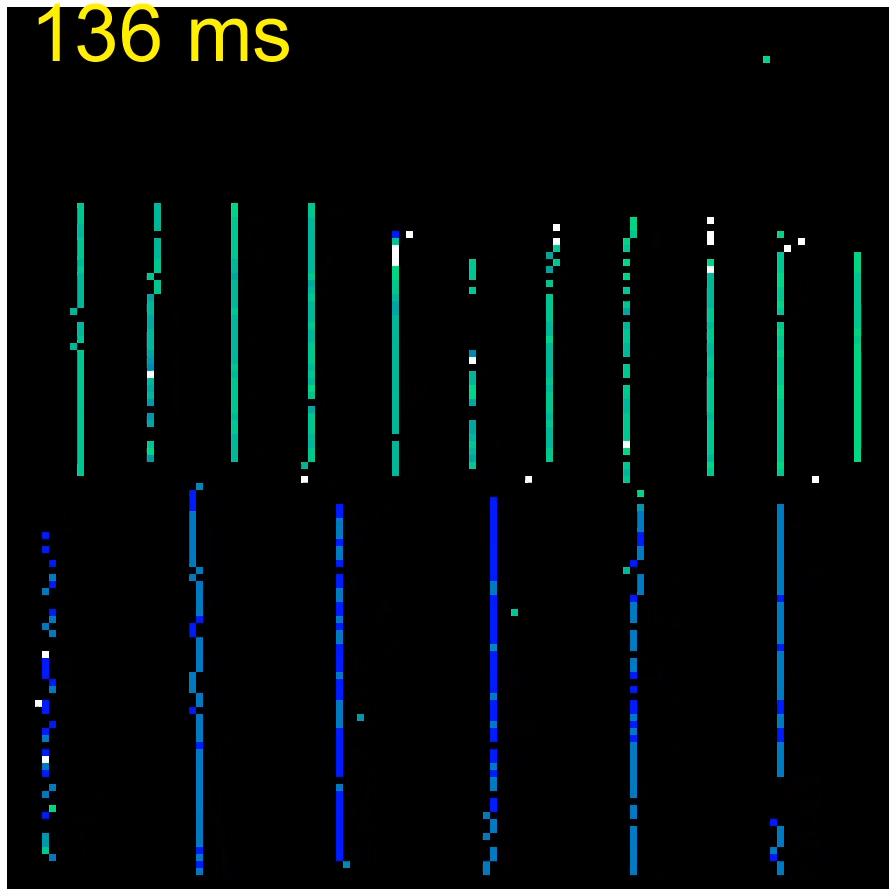}
        \includegraphics[width=0.25\linewidth]{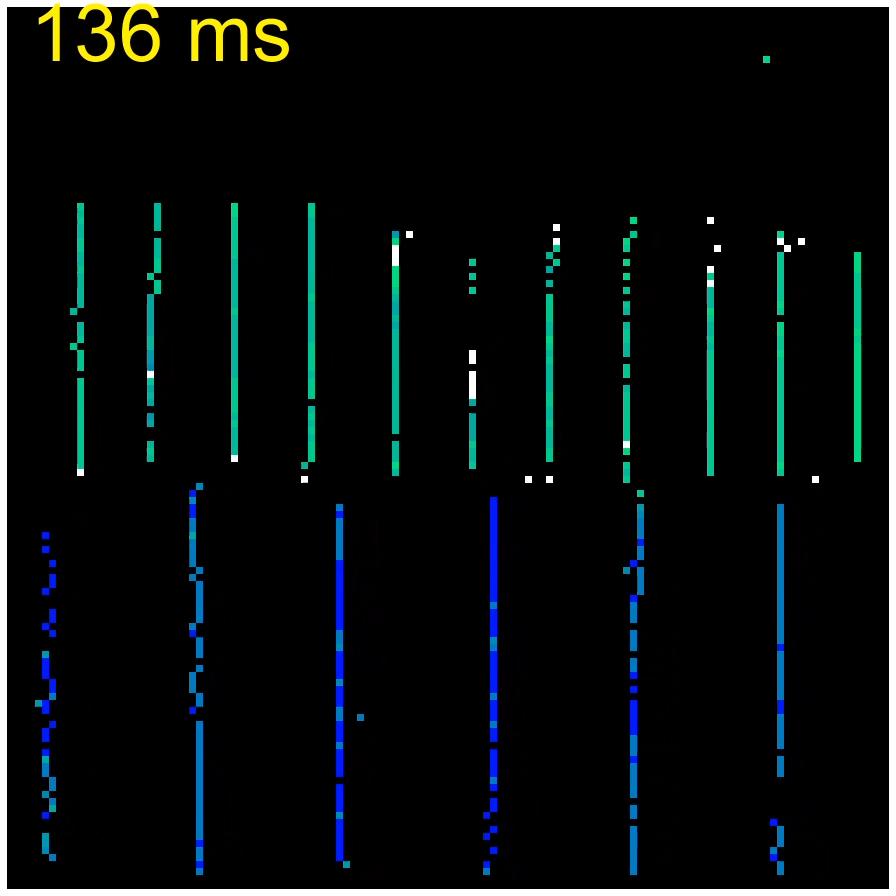}
        \includegraphics[width=0.25\linewidth]{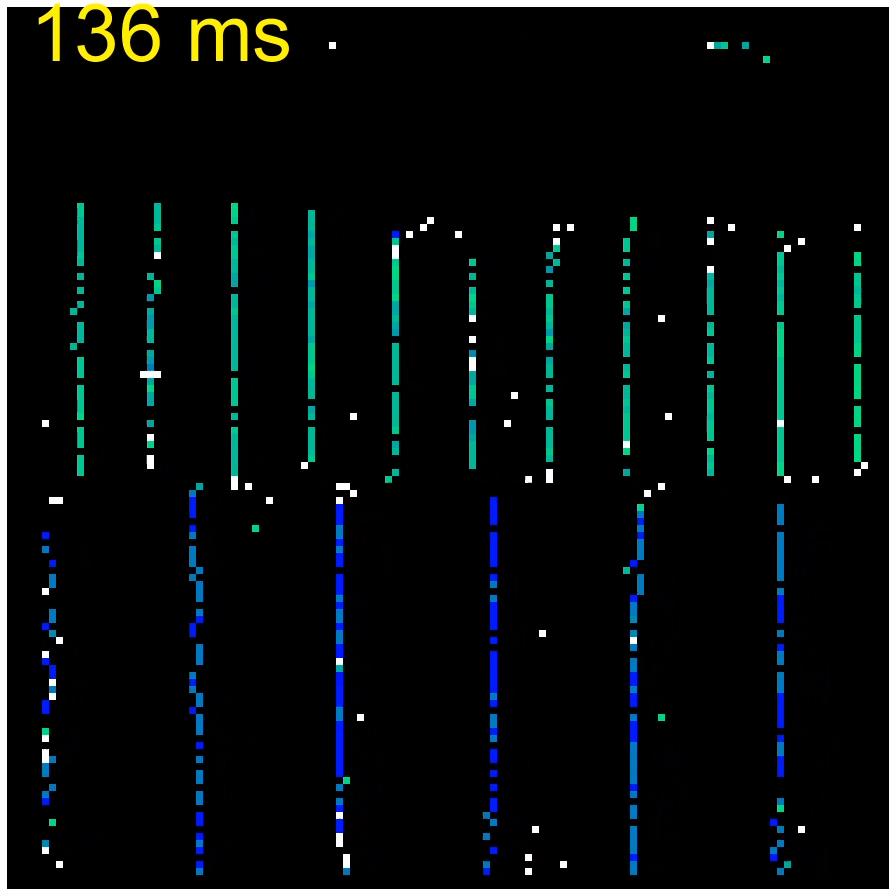}
        
    }
    \caption{From left to right, Events stacked during $20ms$ for the sequence $\tt{stripes}$, active events only shown according to PCA only algorithm, weighted PCA algorithm and leveled PCA algorithm all at $136 ms$.}
\end{figure*}
\textbf{Translation Scenario:} The choice of neighborhood size affects deeply the accuracy of estimating the orientation, for which reason, weighted PCA regularization provided best results. Weighted PCA gives best results because it uses a smaller level of neighborhood for regularization and also because it uses a consensus of neighborhood optical flow which creates a smoothing effect. Leveled PCA regularization improved the estimation of orientation and boosted the accuracy but did not maintain maximum accuracy due to using many levels of the neighborhood.\\

\noindent
\textbf{Rotational Scenario:} As expected, rotational scenarios are harder to estimate and accuracy may be reduced. Results provided by the three methods of PCA did not much vary, however PCA with weightings regularization attained the best results.
\subsection{Lifetime Estimation}
Lifetime is considered as one of the important features assigned to event-based cameras. It helps sharpening image-like matrices used in event-based visual odometry and SLAM algorithms \cite{mueggler}. With sufficiently acceptable accuracy, all PCA algorithms provided reliable lifetime estimation compared to the state-of-the-art \cite{mueggler,Low_2020_CVPR_Workshops}. We used the sequence $\tt{stripes}$ where two stripes are moving at different constant speed and hense constant lifetime for each event generated by these stripes. Figure \ref{fig:hist} shows the probability density function (PDF) of the estimated lifetime. Using PCA only, $52.19\%$ of events where assigned lifetime of $5.35\;ms$ with error of $10.83\%$ of the slow stripes and $17.95\%$ are assigned $10.65\;ms$ with error of $11.25\%$ of the fast stripes. Weighted PCA gave the thinnest distribution around the maximum bins value (least variance). The nearest large bin to slow stripes has the value of $5.55\;ms$ with $57.72\%$ of all events and $7.5\%$ error. For Fast stripes, the largest bin's lifetime is $11.05\;ms$ with $19.44\%$of all events which translates to an error of $7.91\%$ of the actual lifetime. Leveled PCA gives the best results but with the largest variance around actual lifetime with percentage of $38.58\%$ and $15.11\%$ around the slowest and the fastest stripes respectively and errors of $5.83\%$ and $4.58\%$. The high variance of Leveled PCA creates many falsely estimated lifetimes (see Figure \ref{fig:lifetimes}).

\subsection{Computational Time}
PCA method is based on computing the eigen vectors of a $3 \times 3$ matrix that is constructed by increments based on the chosen neighborhood. This results in a much lower computation time compared to other techniques. For the same reason, the change of the neighborhood size, on the contrary to other methods, does not affect the computation time. PCA method only is shown to be 5 times faster than the local plane fitting method and 10 times faster than Lucas-Kanade method. As expected, the usage of levels regularization, which is the slowest PCA method, increased the computation time, however it is still about 2 times faster than local plane fitting methods and about 4 times faster than Lucas-Kanade method. Although these algorithms are implemented on a Linux machine with Core i5 3.10 GHz processor, the provided computation time of PCA variants is shown to be competitive to be selected for real-time applications if implemented using low-level language. 

\section{Conclusion and Future Work}
\label{section:conclusion}
In this paper, we propose a novel method for event-based optical flow estimation based on PCA. We have shown that the proposed method is more adapted to the sparse nature of event-based cameras, and produces significantly less noise than other methods. Although improvements in performance always come at cost of computational power, the method we proposed makes sparing use of the processor and drastically reduces computation time. The simple and straightforward procedure of PCA event-optical flow approach, besides providing accurate and rapidly computed estimation, is the strength  point of the method. We have shown improvements in event-based optical flow estimation and these results incite to pursue the study on event-based optical flow performance for more complex schemes.\\

\noindent
Visual odometry is one of the most critical robotic applications that require the accuracy and agility of the provided data. Based on neuromorphic vision sensors nature and the accomplished gain, a pragmatic decision would be to consider an optical-flow-based visual odometry scheme. To the best of our knowledge, this area is not widely investigated and still requires due attention. To exploit event-based optical flow in robotic applications, a better understanding of its limitations is required. We intend to analyze the capacity of these optical-flow-based schemes in future work.

\end{document}